\documentclass[conference]{IEEEtran}
\IEEEoverridecommandlockouts
\usepackage{amsmath,amssymb,amsfonts}
\usepackage[italic]{mathastext}
\usepackage{algorithmic}
\usepackage{graphicx}
\usepackage{textcomp}
\usepackage{xcolor}

\usepackage[colorinlistoftodos]{todonotes}
\setuptodonotes{inline}
\usepackage[numbers]{natbib}
\usepackage{balance}
\usepackage{hyperref}
\usepackage{booktabs} 
\usepackage{threeparttable}
\usepackage{tabularx}

\def\BibTeX{{\rm B\kern-.05em{\sc i\kern-.025em b}\kern-.08em
    T\kern-.1667em\lower.7ex\hbox{E}\kern-.125emX}}
\begin{document}

\title{Exploring the Use of Robots for Diary Studies\\
}


\author{\IEEEauthorblockN{Michael F. Xu}
\IEEEauthorblockA{\textit{Department of Computer Sciences} \\
\textit{University of Wisconsin--Madison}\\
Madison, Wisconsin, USA \\
\href{mailto:michaelfxu@cs.wisc.edu}{michaelfxu@cs.wisc.edu}\\
\href{https://orcid.org/0009-0000-2632-2428}{0009-0000-2632-2428}}
\and
\IEEEauthorblockN{Bilge Mutlu}
\IEEEauthorblockA{\textit{Department of Computer Sciences} \\
\textit{University of Wisconsin--Madison}\\
Madison, Wisconsin, USA \\
\href{mailto:bilge@cs.wisc.edu}{bilge@cs.wisc.edu}\\
\href{https://orcid.org/0000-0002-9456-1495}{0000-0002-9456-1495}}
}

\maketitle

\begin{abstract}
  As interest in studying in-the-wild human-robot interaction grows, there is a need for methods to collect data over time and in naturalistic or potentially private environments. HRI researchers have increasingly used the \textit{diary method} for these studies, asking study participants to self-administer a structured data collection instrument, \textit{i.e.}, a \textit{diary}, over a period of time. Although the diary method offers a unique window into settings that researchers may not have access to, they also lack the interactivity and probing that interview-based methods offer. In this paper, we explore a novel data collection method in which a robot plays the role of an interactive diary. We developed the \textit{Diary Robot} system and performed in-home deployments for a week to evaluate the feasibility and effectiveness of this approach. Using traditional text-based and audio-based diaries as benchmarks, we found that robots are able to effectively elicit the intended information. 
  We reflect on our findings, and describe scenarios where the utilization of robots in diary studies as a data collection instrument may be especially applicable.
\end{abstract}


\section{Introduction}

The \textit{diary method} involves participants self-reporting their life experiences \textit{in situ} on a regular basis with the use of a ``diary'' \cite{nezlek2012diary}. Compared to other research methods, diary methods allow researchers to examine the experiences of participants in natural and spontaneous settings. They also reduce the level of retrospection by minimizing the time lapse between an experience and the account of that experience~\cite{bolger2003diary}.
The diary method has been used across fields ranging from nursing research \cite{ten2018voice} to studying organizations \cite{ohly2010diary}. Although the diary method originated from the fields of psychology, anthropology, and history to capture ``life as it is lived'' \cite{bolger2003diary}, it has since been adopted in human-computer interaction (HCI) research to study the use of or inform the design of technology \cite{carter2005participants}. HCI researchers have utilized the diary method to investigate a variety of topics, including agency delegation in the context of meal plan services~\cite{verame2018learning}, limits on and transitions from young children's screen time~\cite{hiniker2016screen}, and privacy perceptions and concerns with respect to smart speakers~\cite{lau2018alexa}.

\begin{figure}[t]
  \centering
  \includegraphics[width=\linewidth]{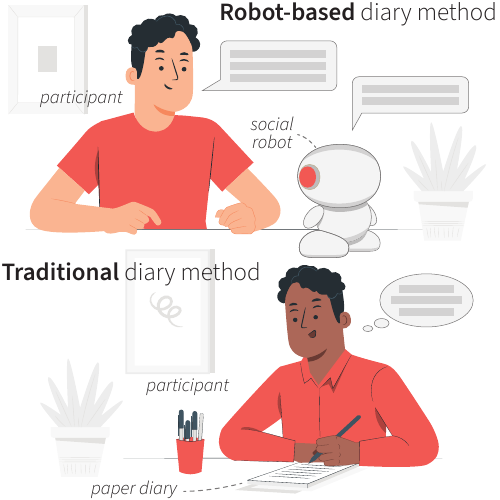}
  \caption{In this paper, we explore the use of conversational social robots as instruments for data capture in diary studies, as an alternative to traditional diary methods that use paper, audio, or computer-based data capture.} 
  \label{fig:teaser}
  \vspace{-18px}
\end{figure}

The increasing interest in studying in-the-wild human-robot interactions highlights a need for methods that collect longitudinal data in naturalistic or private environments without the presence of researchers. Several recent HRI studies have adopted the diary method. For example, in a deployment of ``Joy for All'' dog and cat robots in assisted living facilities, researchers asked caregivers to log interactions between older adults and the robots~\cite{bradwell2020longitudinal}. In an in-home ``robot camp'' deployment, parents and children together recorded their co-learning experience with a social robot through an online diary~\cite{ahtinen2023robocamp}. In a study exploring domestic robots specialized in task automation, users were asked to keep a diary during the 10-day deployment~\cite{schneiders2021domestic}. 
%
Researchers have also used the diary method to collect data during predeployment stages. For example, to inform the design of a robot that helps workers manage taking breaks throughout the day, \citet{vsabanovic2014designing} conducted a diary study in which participants recorded the time and reasons for their breaks, and developed prototypes based on these insights. Separately, in an attempt to identify and characterize \textit{opportune moments} for the robot to interact with their users, \citet{hsu2024now} performed a diary study that involved sending users text messages over a two-week period. 
These applications of the diary method highlight its advantage in collecting data in various naturalistic or private settings that are often relevant to in-the-wild HRI studies.

In light of the increasing need for the use of diary study methods in HRI, this study proposes and evaluates a novel diary-based data collection method: the use of a robot as an interactive diary. 
Although the diary method enables data collection in scenarios and settings where traditional survey- or interview-based methods are ineffective or infeasible, this method still lacks the interactivity and probing that interview-based methods offer. Robots, on the other hand, have shown promise in facilitating interactive data collection through interviews \cite{briggs2015robots, bethel2016using, boumans2019robot, laban2024building}. 

In this work, we leverage the potential of robots to effectively conduct interviews, and develop the \textit{Diary Robot} system, where a conversational robot plays the role of an interactive diary. We conducted in-home diary studies and asked participants to document daily routines, focusing on their children's bedtime routines, for a week through conversations with the robot. The choice of children’s bedtime routines as the topic was twofold. First, it embodies the naturalistic and private characteristics we highlighted as being particularly suitable for diary studies. Secondly, it aligns with our own research into long-term HRI, where we investigate family routines and how robots can support them~\cite{xu2024robots, cagiltay2024toward}, hence demonstrating its applicability in authentic HRI research. To evaluate this method in a between-subject study design, we compared the conversational robot with the use of text-based and audio recording-based diaries. 
Through analysis of diary entries and the post-study questionnaire and interviews, we found that robots are able to effectively elicit the intended information. 

The contributions of this work are three-fold:
\begin{itemize}
  \item \textit{System Contributions:} We present the Diary Robot system, a versatile framework that could be applied to various platforms, and adapted to conduct diary studies on various topics.
  \item \textit{Empirical Contributions:} We compare the characteristics of the data collected by the three formats, and evaluate the effectiveness of using robots as interactive diaries.
  \item \textit{Design Implications:} We reflect on our findings, and discuss considerations for utilizing this system as a data collection method in HRI. 
\end{itemize}

\section{Related Work}

\subsection{Diary Study Methods}
Diary methods have been implemented in various formats. Building on traditional paper and pen diaries, some digitized variations that have been utilized include online questionnaires~\cite{hiniker2016screen, lau2018alexa}, entries created in shared online workspaces~\cite{overdevest2023towards, ahtinen2023robocamp}, and emails or text messages sent directly to researchers~\cite{sung2009robots, hsu2024now}. In the early 2000s, \citet{palen2002voice} experimented with voicemails as a new diary creation interface. Over the years, researchers have also explored enriching diary entries through the inclusion of audio recordings~\cite{williamson2015evaluating, worth2009making} and captured photos~\cite{verame2018learning}.

Each of these methods has its own pros and cons. As discussed in \citet{bolger2003diary}, traditional paper and pen diaries are simple to implement and have a low technical barrier for users. However, the drawbacks are also salient: the burden is usually high on the participants, mistakes and compliance are hard to monitor or correct, and data entry is error-prone and cumbersome. Digitized versions of this method mitigate many of the drawbacks, but text-based input remains a burden on participants. For example, \citet{overdevest2023towards} reported that even when using shared online workspaces, their participants were ``reluctant to fully fill out duplicate diary entries'' when they perceive the content to vary little between entries. Audio diaries mitigate the burden of writing or typing, but raise new challenges. In their evaluation of audio diary methods, \citet{williamson2015evaluating} found that their participants reported mixed experiences, some finding it therapeutic, while others raised concerns about confidentiality and increased self-awareness. In another study on audio diaries, the researchers recounted that one of their participants found it strange talking to a recorder, and had to picture a listener in their mind to alleviate the feeling~\cite{worth2009making}. Similarly, \citet{hyers2018diary} described audio diaries as an exciting format, but also warned of the possibility of ``longer, rambling sentences and more complex speech.'' This increases the challenge for researchers to analyze the data, and potentially causes the users to lose focus on the important topics.



\subsection{Data Collection through Human-Robot Conversations}
Prior work has extensively studied the ways in which virtual characters and robots can play a role in the elicitation of self-disclosure \cite{powers2007comparing, lucas2014s, martelaro2016tell, akiyoshi2021robot, li2023tell, laban2024building}. For example, in a study on self-disclosure in mental health, \citet{li2023tell} found that some participants reported an appreciation for the judgment-free space to disclose their feelings when they interacted with a virtual robot,
while others found the interaction with the virtual robot ``creepy'' and ``awkward.'' Compared to user experiences with writing and solo speaking, they also found that participants preferred interactions with the robot as the platform of self-disclosure.

Other studies investigated robot's potential in collecting data in interviews. In the context of collecting health status of subjects with Parkinson's disease, \citet{briggs2015robots} compared user experiences between interactions with a human interviewer and a Nao robot interviewer. They found that while participants generally viewed robot interviewers positively, they still preferred human interviewers. Furthermore, \citet{boumans2019robot} explored the use of Pepper robots to interview older adults about their health data. Although they concluded that social robots can be effectively used in these scenarios, \citet{stommel2022pepper} later found an abundance of miscommunication within the data, particularly due to the robot mishearing and causing participants to repeat and rephrase themselves. Robot interviewers have also been used to investigate bullying among children~\cite{bethel2016using}, conducting in-person satisfaction surveys~\cite{heath2020challenges}, etc. These studies highlight the potential for robots as an interactive data collection instrument, but also caution researchers about the various potential failure points.

In this paper, we present the \textit{Diary Robot} system as an embodied and interactive data collection instrument for diary studies, intended to take advantage of the interviewing capabilities of robots, and mitigate some of the weaknesses of existing diary formats. To address challenges with robot interviewers, we adopted advanced speech to text (STT) and large language models (LLMs), and designed the interactions to be highly structured.
The next section discusses the design and implementation of our system.


\section{The Diary Robot System}
\label{sec:system}
The \textit{Diary Robot} system consists of a Misty II robot, a Raspberry Pi unit (the controller), and optional accessory devices for networking setup. In this setup, the Misty robot is predominantly utilized as a puppet device, where the interactions are all orchestrated by the controller. Communications between the robot and the controller happens through a router, either the participant's home router or one that we provide. 


At a high level, the system could be in three states: \textit{Idling}, \textit{Chatting}, or \textit{Diary} mode. From Idling state, users can initiate Chat mode and engage in free-form chatting with the robot, by pressing the bumper of the robot. While engaging with the robot in free-form conversations, users may then transition to the Diary mode by verbally saying something along the lines of ``Can we start the diary entry activity?'' To enable free-form conversations, we used various OpenAI APIs. After audio was captured by the robot and transferred to the controller, we converted it to text using \textit{Whisper}, generated replies using GPT (\textit{gpt-4o-2024-05-13}) and then converted the responses to audio format using OpenAI's \textit{alloy} voice. Auto-detection of end-of-speech, a capability built into the Misty II robot, was also utilized for the free-form conversations.

\begin{figure*}[tbp]
  \centering
  \includegraphics[width=\linewidth]{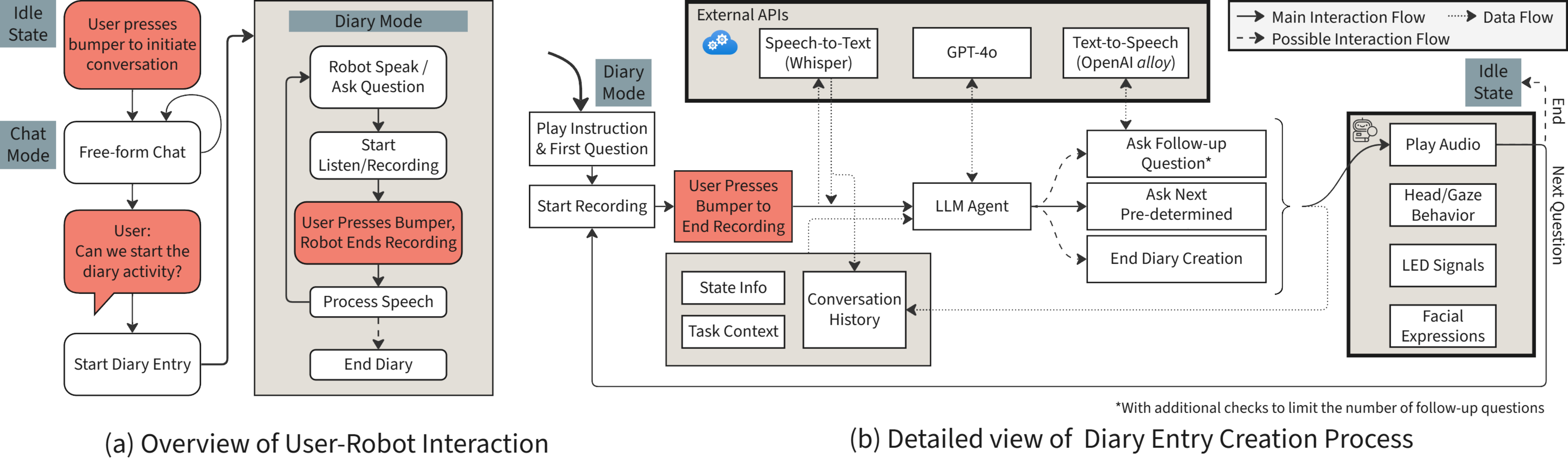}
  \caption{Illustrations of the Software Architecture. Once the user initiates free-form chat, they can verbally command the robot to start a diary entry. In the diary entry mode, the system follows a structured flow, going through a set of predetermine questions about bedtime routines. After each predetermined question, possible follow-up questions may be asked, determined by the embedded LLM-agent. All interactions (playing and recording, non-verbal expressions, etc.) are performed through the robot, and all the processing and external API requests are handled by the controller.}
  \label{fig:software_diagram}
  \vspace{-12px}
\end{figure*}

Once in the Diary mode, the system follows a structured flow, where it goes through a set of six predetermined questions about the participant's child's bedtime routine that night (the topic we selected for the diary entries). After each question, possible follow-up questions may be asked, determined by the embedded LLM-agent powered by \textit{gpt-4o}. Hard-coded limits are also in place to cap the number of follow-ups for each questions. The follow-up questions serve several purposes, as they increase the interactivity, allow the system to recover from incorrect STT processing, and enable participants to supplement earlier answers.
In the Diary mode, to allow the user ample time to respond, we disabled the auto-detection of end-of-speech, and instructed the participants to manually press a bumper to indicate the end of a response. An overview of the software architecture is illustrated in Fig.~\ref{fig:software_diagram}. The structure of the prompt we specified for the GPT requests are provided as supplementary materials.\footnote{\url{https://osf.io/5hya6/}} 

We also implemented various non-verbal modalities of communication. For example, LED lights were used to signal various states of the system: glowing green on the chest indicates \textit{ready for interaction} and blue on the side of its head indicate that it is \textit{listening (recording)}. Basic facial expressions were programmed to match the state of the interaction, such as a \textit{pensive} expression when the robot is processing the input. To improve the naturalness of the conversational interactions, we also implemented gaze behaviors (through head movements) based on \citet{andrist2014conversational}.

\section{Method}
\label{sec:method}

\subsection{Diary Study}
To investigate the feasibility and effectiveness of utilizing a robot as a diary data collection instrument, we developed and deployed the \textit{Diary Robot} system to local families recruited through a university staff mailing list. During the study, and as the topic of the proposed diary study, participants were asked to document their child's bedtime routine for seven consecutive days. The following study design was approved by the author's Institutional Review Board.

\subsection{Conditions} 
To better understand the strengths and weaknesses of using a robot as an interactive diary, and compare it against existing diary formats, we designed and implemented three conditions. Participants in the \textit{Robot} condition created the diary entries through conversations with the \textit{Diary Robot}. Participants in both the \textit{Text} and \textit{Audio} conditions created their entries on an online questionnaire, with the former presented with text boxes for input, and the latter submitting their input through audio recordings.


\subsection{Participants}
We recruited parents who had at least one child aged between five and twelve. Specifically, the inclusion criteria were: Fluent in English, 18 years or older, has at least one child aged 5-12 living at home, and within one hour driving distance from the author's university campus. The main exclusion criterion was cognitive or physical disability that would prevent the parent from participating in the study conditions.
24 parents participated in our study. 19 participants self-reported themselves as the mother of their family, and the majority of the families are considered middle- to upper-income households (\textit{i.e.}, self-reported annual household incomes of \$100,000 USD or higher). We refer to the participants as R1--R8, T1--T8, and A1--A8, where the prefix indicates the condition the participant was assigned to (\textit{Robot}, \textit{Text}, and \textit{Audio}).

    


    



\subsection{Study Design}
We conducted a between-subject study, randomly assigning participants to one of the three conditions. For each participant, the study consisted of three sessions: The \textit{setup session}, where the experimenter performed an in-person visit to introduce the study, and set up the equipment for the \textit{Robot} condition. The \textit{main study} period, where the participants complete the week-long diary study, performing daily documentation of their children's bedtime routines. The \textit{post-study session}, where the experimenter conducted a post-study questionnaire and semi-structured interview. 


During the main study, six predetermined questions about the bedtime routines were asked every night, and the questions were the same across the three conditions. Some examples of the questions include: \textit{``What are the steps involved in your child's bedtime routine, how did they go tonight;''} \textit{``Were there any challenges during tonight's bedtime routine;''} and \textit{``How were you feeling by the end of the routine? Why did you feel that way?''} The full list of predetermined questions are provided in the supplementary materials.





\subsection{Study Procedure}

All setup and post-study sessions were conducted in-person.
%
During the \textbf{setup session}, we provided the participants with an information pamphlet, including instructions on when and how to create a diary entry. The experimenter reviewed the pamphlet and demonstrated how to create a diary entry with the participants. 
The participant and the experimenter also agreed on a reminder schedule, describing the participant's preference on 1. \textit{When} the experimenter checks their diary entry submission each night, and 2. \textit{how} the experimenter sends the reminders (\textit{i.e.}, text messages or emails). Participants were encouraged to create entries for all seven days of their main study, though only five were required. We informed participants that if circumstances do not allow, creating an entry the next day is also acceptable, but diary entries created more than one day after the intended day will not be considered valid by our protocol. For the \textit{Robot} condition, this session also included setting up the \textit{Diary Robot} system. In general, this session was around 10 minutes for the \textit{Audio} and the \textit{Text} conditions, and around 40 minutes for the \textit{Robot} condition.

During the \textit{main study}, each night, the experimenter monitored the diary entry creations, and sent out reminders as needed, following the agreed upon schedule.
During the \textit{post-study session}, participants first completed a questionnaire on a tablet, followed by a semi-structured interview about their experience creating the diary entries. All participants were compensated \$50 USD for completing the study.

\subsection{Data Collection and Analysis}
We conducted a mixed-method analysis on four types of data collected: the diary entries created by the participants, the questionnaire responses from the post-study session, the post-study semi-structured interviews, and compliance data. We used the robot's system logs to contextualize the data.  

\subsubsection{Diary Entries}
\label{sec:data:diary_entries}
All diary entries created during the week-long field deployment were collected digitally. For both the \textit{Text} condition, and the \textit{Audio} condition, participants completed an online questionnaire hosted on Qualtrics~\cite{qualtrics}. Instead of seeing a text box as the input format as the participants in the \textit{Text} condition would, participants in the \textit{Audio} condition were provided buttons to record their audio responses. 
For the \textit{Robot} condition, data was saved to the device and collected post-study after the retrieval of the system kits. 

Following existing literature \cite{andrist2014conversational, moon2000intimate}, we utilized the word count as a first approximation of disclosure and information richness. We define a \textit{diary entry} to be the concatenation of all the responses to the questions on a single day. We calculated the word count for each entry, as well as the average word count per entry for each participant, and analyzed the distributions for the three conditions.

To develop a criterion to quantify and evaluate the extent to which the diary entries supported the goal of the diary study (\textit{i.e.}, understanding children's bedtime routines in families), we performed a content analysis on the diary entries. Specifically, we analyzed the following \textit{seven dimensions}. \textit{Feelings \& Thoughts} includes codes describing the participant's own mental state such as ``felt good,'' ``depleted,'' and ``frustrated.'' \textit{Bedtime Activities} covers the main steps and actions involves in the bedtime routine (\textit{e.g.}, brush teeth, and put on pajamas). \textit{Other Activities} refers to other actions and activities that were discussed but not as part of the bedtime routine (\textit{e.g.}, art project, exercise, movie, and piano). \textit{Child Remark} covers thoughts and comments the participant had about their child (\textit{e.g.}, exhausted, independent, and good listener). \textit{Other Details} covers other topics or details that were not captured by the other dimensions (\textit{e.g.}, the conversation topic, and the participant's own plans \& desires for the night). We also looked at two numerical dimensions. \textit{Reasons Given} counted the number of reasoning or goal of an action, emotion, or comment that was mentioned (\textit{e.g.}, ``[child] finished dinner a little bit before [us], so she was on her tablet a little bit before bedtime'' would count as one). Finally, we also counted the number of timing related instances mentioned (\textit{e.g.}, ``8pm'' and ``10 minutes'' would each count as one instance).
We iteratively developed a code book, and the first author coded all the diary entries. A separate coder who was blinded to the study independently coded a subset of the diary entries. We report aggregated inter-rater reliability scores ranging between $0.8$ and $0.95$ for the five categorical dimensions, with an overall value of $0.88$. The Spearman rank-order correlation coefficients for the number of reasons and timings given, are approximately $0.8$ and $0.9$, respectively.


For each of these dimensions, we counted the number of instances in each diary entry, and aggregated the counts at the participant level (\textit{i.e.}, summed across the multiple diary entries created by the same participant). We analyzed both total counts and unique counts, where the latter only counts the same type of information once even if it was mentioned multiple times.

Finally, as a comparison to the word-count analysis above, we defined the \textit{overall information} to be the sum of all the (unique) counts of the seven dimensions, as an approximation to how much we were able to learn about each participant through their diary entries over all of their entries. We then performed the same analysis as we did for the word count, examining the trends and averages.


\subsubsection{Post-study Questionnaire}
The questionnaire consisted of three main sections: (1) the Usefulness, Satisfaction, and Ease of use questionnaire (USE) \cite{lund2001measuring}; (2) the System Usability Scale (SUS) \cite{brooke1996sus}; and (3) a modified subset of questions on subjective self-disclosures, adopted from \citet{parks1996making}. For subjective self-disclosure, we adapted items from the \textit{Breadth} and \textit{Depth} section of \citet{parks1996making}, and modified them to be more applicable to our study context. For example, an item from the \textit{Depth} section ``I usually tell this person exactly how I feel'' was updated to ``I usually document in the diary entries exactly how I feel.'' In the end, there were three questions in the breadth category and eight questions in the depth category. The full list of questions are provided in the supplementary materials. All questions are based on anchored Likert scales, where the options ranged from ``1 (Strongly disagree)'' to ``7 (Strongly agree)'' for both questions in USE and questions for the subjective self-disclosure, and ``1 (Strongly disagree)'' to ``5 (Strongly agree)'' for questions in SUS. We conducted a confirmatory analysis on the internal consistencies of the responses. For the USE dimensions and the SUS responses, this analysis reassured us with Cronbach’s alpha values between $0.847$ and $0.966$. However, Cronbach’s alpha for the self-reported breadth and depth were $0.420$ and $0.732$, respectively. This low reliability of the breadth scale is likely due to the selective usage of the original scale (three out of the five items) and the wording changes applied to adapt them to an HRI context. Therefore, in the analyses that ensue, we dropped one question that may have been made too complicated for the participants to process, and kept the remaining two breadth related questions as separate dimensions: \textit{Scope} (``go beyond what we were asked to share'') and \textit{Flow} (``move easily from one topic to another'').

\subsubsection{Post-study Interview}
The semi-structured interview included questions about the participant's experience, factors influencing their on-time diary entry creations, concerns about privacy, and so on. We performed a reflexive thematic analysis  \cite{braun2006using, byrne2022worked, braun2019reflecting} on the interview transcripts, with a focus on contextualizing the results from other sources.

\subsubsection{Compliance}
Compliance data were logged manually, categorized as either on-time (with \textit{vs.} without a reminder), late (submitted on the next day), or missed (no submission or more than one day late).

In total, the 24 participants created 162 valid diary entries. For the various quantitative measures, we computed descriptive statistics and performed pairwise comparisons among the three conditions using Tukey’s honestly significant difference (HSD) test. All Tukey’s HSD tests were performed at the participant-level ($N=8$ for each condition).



\section{Results}
We report results on (1) the information richness of the diary entries based on word count and our content analysis; (2) the system's usability based on the questionnaire and interviews; (3) the factors affecting user experiences based on interviews and system logs; and (4) participant compliance. Statistical results on selective measures are reported in Table~\ref{tab:stats}, and all other test statistics are provided in supplementary materials.

\subsection{Information Richness and Disclosure}
We analyzed the diary entries created by participants and their subjective measures reported in the questionnaire. 
On average, participants in the \textit{Robot} condition generated diary entries with comparable information to the \textit{Audio} condition, but included more bedtime routines than the \textit{Text} condition.


\subsubsection{Word Count}
On average, entries from the \textit{Audio} condition had the highest word count ($M=394$; $SD=299$), followed by the \textit{Robot} condition ($M=286$; $SD=166$), and then the \textit{Text} condition ($M=130$; $SD=30.9$). In Fig.~\ref{fig:wc_info_overtime}~$(a)$, for each participant, we present their word count over the week-long study, grouped by their conditions. In Fig.~\ref{fig:disclosure_distributions}~$(a)$, we calculated the average word count for each participant and present the distribution by conditions. Although entries from the \textit{Robot} condition on average contained more than double the word count of those from the \textit{Text} condition, the only statistically significant difference were between the \textit{Audio} and the \textit{Text} conditions, at $p=0.037$. 


\subsubsection{Information Richness by Topics and Categories}
\begin{figure}[tbp]
  \centering
  \includegraphics[width=\linewidth]{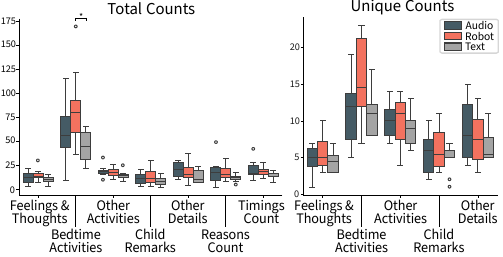}
  \caption{Total and Unique number of occurrences in diary entries, aggregated at the participant-level. $**$ denotes $p<0.05$, and $*$ denotes $p<0.1$.}
  \label{fig:topic_count}
\end{figure}

We analyzed the diary entries for their information richness by each of the seven dimensions described in Section~\ref{sec:data:diary_entries}, and present the findings in Fig.~\ref{fig:topic_count}. In the panel on the left, we show the total number of times that information has been provided for each dimension (\textit{e.g.}, \textit{Feelings \& Thoughts}) over the course of the study for each participant. On the right, we show the distribution for \textit{unique} counts, where multiple mentions of the same information were only counted once.

\begin{figure}[tbp]
  \centering
  \includegraphics[width=\linewidth]{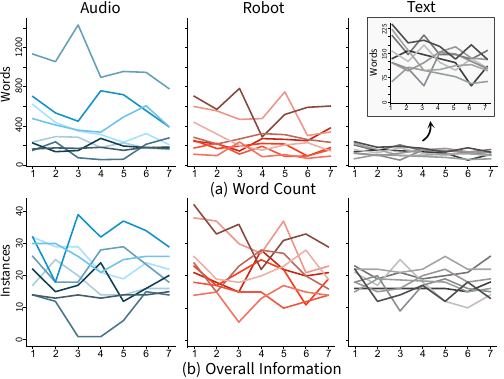}
  \caption{Word count and overall information over the course of the week.}
  \label{fig:wc_info_overtime}
\end{figure}

\begin{figure}[bp]
  \centering
  \vspace{-12pt}
  \includegraphics[width=\linewidth]{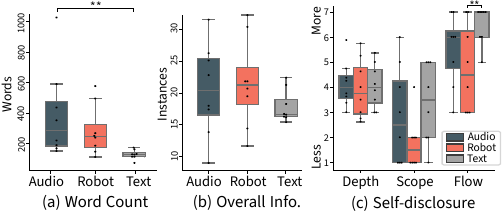}
  \caption{Distribution of average word count, overall information, and subjective self-disclosures. $**$ denotes $p<0.05$, and $*$ denotes $p<0.1$.}
  \label{fig:disclosure_distributions}
\end{figure}

We observed that on average, participants in the \textit{Robot} condition reported the highest number of total bedtime routine activities ($M=84.9$; $SD=43.1$). Participants in the \textit{Audio} and \textit{Text} conditions reported $59.9$ ($SD=31.9$) and $M=44.8$ ($SD=17.4$), respectively. The difference between the \textit{Robot} condition and the \textit{Text} condition was marginal, $p=0.056$. The differences between the other two pairs were not significant; although we observed a trend where participants in the \textit{Robot} condition reported $42\%$ more bedtime routine activities than those in the \textit{Audio} condition, more data is required to establish statistically meaningful differences.


In Fig.~\ref{fig:wc_info_overtime}~$(b)$ and Fig.~\ref{fig:disclosure_distributions}~$(b)$, we summarized the level of \textit{overall information} contained within the diary entries. In contrast to the findings for word counts, we do not observe any statistically significant differences among the three conditions. Specifically, the \textit{Robot} condition ($M=21.6$; $SD=7.0$) on average contained a similar level of overall information compared to the \textit{Audio} condition ($M=20.5$; $SD=7.32$).


\subsubsection{Subjective Self-Disclosure}
Although participants from the three conditions on average reported similar levels of self-disclosure in terms of depth, participants from the \textit{Robot} condition reported lower breadth in terms of the \textit{flow} dimension (ease of moving between topics) ($M=4.75$; $SD=1.75$), compared to the \textit{Text} condition ($M=6.50$; $SD=0.76$, $p=0.047$.) For the \textit{scope} dimension of breadth, seven of the eight participants from the \textit{Robot} condition responded with either a one or a two (Fig.~\ref{fig:disclosure_distributions}~$(c)$). This pattern is consistent with what we observed in the analysis of the diary entry content (Fig.~\ref{fig:topic_count}): the \textit{Robot} condition covers comparable amount of information on dimensions like \textit{Other Details} but exhibits a trend to include higher counts for activities directly related to the bedtime routine.

\begin{figure}[bp]
  \centering
  \vspace{-12pt}
  \includegraphics[width=\linewidth]{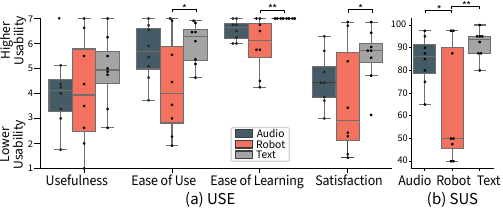}
  \caption{Distribution of Responses to USE and SUS. $**$ denotes $p<0.05$, and $*$ denotes $p<0.1$.}
  \label{fig:post_study}
\end{figure}

\subsection{Usability, Ease of Learning, and Satisfaction}
\label{sec:usability}
On average, participants in the \textit{Text} condition reported the highest rating, while participants in the \textit{Robot} condition reported the lowest. This pattern is consistent across the various sub-scales. We present the distribution of the responses to USE in Fig.~\ref{fig:post_study}~$(a)$, and SUS in Fig.~\ref{fig:post_study}~$(b)$. For example, the average SUS value for the \textit{Robot} condition was $63.8$ ($SD=25.7$), compared to $84.4$ ($SD=10.8$) for the \textit{Audio} condition, and $91.6$ ($SD=6.26$) for the \textit{Text} condition. The differences between the \textit{Robot} and \textit{Text} condition were either significantly or marginally different except for the \textit{Usefulness} subscale in USE. The differences between the \textit{Robot} and \textit{Audio} condition were not statistically significant, except for the SUS, where the difference was marginal. This pattern where the online questionnaires are perceived to be easier to use is not unexpected, especially with our participant population, for their general familiarity with computer-based information capture. In fact, multiple participants (\textit{e.g.}, T4, T8, A5) explicitly mentioned being familiar with the survey platform, Qualtrics. We also observed a bimodal distribution unique to the \textit{Robot} condition. Specifically, there is a higher level of variance in terms of user experiences within the \textit{Robot} condition, where one subgroup rated the system highly across the sub-scales, yet the other rated it low, with few participants in-between. This is especially noticeable for the SUS ratings, where three of the eight participants responded with a value of $85$ or higher, while the other five rated it at $50$ or below.

\begin{table*}
  \caption{Descriptive and Inferential Statistics on various measures. Tukey’s honestly significant difference (HSD) test was used to perform pairwise tests. 
  $*$ denotes $p<0.1$, and $**$ denotes $p<0.05$.}
  \label{tab:stats}
  \begin{tabular}{lccccccccccccccccc}
    \toprule
    & \multicolumn{2}{c}{Robot} &&
        \multicolumn{2}{c}{Audio} &&
        \multicolumn{2}{c}{Text} &&
        \multicolumn{2}{c}{R - A} &&
        \multicolumn{2}{c}{R - T} &&
        \multicolumn{2}{c}{A - T}\\
    
    \cline{2-3} \cline{5-6} \cline{8-9} \cline{11-12} \cline{14-15} \cline{17-18}
    
    & M & SD && M & SD && M & SD &&
        \textit{diff} & $p$ && \textit{diff} & $p$ && \textit{diff} & $p$ \\
    
    \midrule

    Word Count & 286 & 166 && 394 & 299 && 130 & 30.9 && 
        -108 & 0.53 && 156 & 0.28 && 264 & 0.037$^{**}$ \\

    \textit{Total} Bedtime Act. & 84.9 & 43.1 && 59.9 & 31.9 && 44.8 & 17.4 &&
        25.0 & 0.30 && 40.1 & 0.056$^*$ && 15.1 & 0.628 \\

    \textit{Unique} Bedtime Act. & 15.5 & 6.00 && 11.4 & 4.87 && 11.0 & 3.16 &&
        4.13 & 0.22 && 4.50 & 0.17 && 0.38 & 0.99 \\

    Overall Information & 21.6 & 7.05 && 20.5 & 7.32 && 17.9 & 2.59 &&
        1.08 & 0.93 && 3.69 & 0.46 && 2.61 & 0.67 \\

    SUS & 63.8 & 25.7 && 84.4 & 10.8 && 91.6 & 6.26 &&
        -20.6 & 0.052$^*$ && -27.8 & 0.008$^{**}$ && -7.19 & 0.66 \\

    Self-disc.: Scope & 1.75 & 1.04 && 2.88 & 1.96 && 3.38 & 1.60 &&
        -1.13 & 0.346 && -1.63 & 0.123 &&  -0.50 & 0.803 \\

    Self-disc.: Flow & 4.75 & 1.75 && 5.50 & 1.41 && 6.50 & 0.76 &&
        -0.750 & 0.582 && -1.75 & 0.047$^{**}$ &&  -1.00 & 0.331 \\

    Self-disc.: Depth & 3.94 & 1.18 && 4.13 & 0.91 && 4.06 & 0.89 &&
        -0.188 & 0.93 && -0.125 & 0.97 &&  0.062 & 0.99 \\
    
  \bottomrule
\end{tabular}
\end{table*}

    
    
    









    

\subsection{Factors Affecting User Experience}
Based on the thematic analysis of the post-study interviews, we summarized five main factors that affected user experiences. First, multiple participants stated that creating diary entries by interacting with a robot was \textit{fun} (R2, R3, R7); that they were \textit{``looking forward to [creating the diary entries] a little bit more''} (R3); and that the Misty robot was \textit{cute}, which, together with other factors such as its facial expressions, made it easier to talk to and have a conversation with (R8). 

Second, this \textit{conversational format} was generally perceived positively by the participants, evoking a sense of closeness (R7), inspiring them to reflect more on their feelings (R4), and allowing them to be more descriptive and include more details in their diaries (R3). Specifically, regarding the conversations, two participants mentioned that the \textit{follow-up questions} played a positive role (R4, R7). R7 stated that using a robot as an interactive diary was \textit{``pretty good because phone recording never gave us follow up questions.''} Additionally, the robot's reflection on topics that the participant had mentioned before offered the participant a sense of being heard and understood.

Third, some participants experienced frustration or were concerned about interacting with the robot. Among the main factors brought up are \textit{technical glitches} and other \textit{system limitations}, such as processing lag during conversations. Contextualizing interview transcripts (\textit{e.g.}, R6 mentioning difficulty in initiating the diary creation) with system logs, we found that some occasional technical glitches included incorrect end-of-speech detection (causing the wrong audio snippet to be processed) and inaccurate speech recognition, both leading the conversation awry at times during the free-form chatting, affecting the transition into the diary activity. 

Fourth, the \textit{time required to start up} the system was mentioned by multiple participants (R1, R2, R4, R8). R4 stated: \textit{``I wish it would've been more speedy, [...] (the start-up time was) like a minute or so, which I guess isn't that terrible but when you're used to iPads and the computers booting up right away, sometimes I would feel frustrated.''} 

Fifth, \textit{privacy} was a concern for many when a robot is placed in the home (R5, R6, R7). During the study, R5 kept all the equipment unplugged and powered off, except when she needed to interact with the \textit{Diary Robot}. She described the reasoning as \textit{``I guess there was just a little bit of wanting to have like just our regular private life.''} This concern about privacy could also have affected the amount and quality of information submitted within the diary entries: R6 stated that she \textit{``kept it very short, [...] not sharing the thing that I would be uncomfortable on being out there.'' }

\subsection{Compliance}
\label{sec:compliance}
Despite the lower self-reported usability metrics, the \textit{Robot} condition performed best in terms of compliance.
The vast majority of entries ($87.5\%$) from the \textit{Robot} condition were submitted on the intended night and without the need for a reminder. The corresponding percentages of on-time submissions without reminders were $71.4\%$ for the \textit{Audio} condition, and $59.0\%$ for the \textit{Text} condition. 
Several factors unique to the \textit{Robot} condition may have contributed to the higher compliance rate. For example, the physical presence of the robot was mentioned a few times (R2, R3, R4). R3 described to us \textit{``...I think having it, something present in the house rather than a piece of paper, something that I look at every day when I walk past it, reminds me to do it and I feel like that was helpful.''} 
Another factor that was not mentioned by participants in either the \textit{Audio} or \textit{Text} condition, is the role of family members' involvement. R5 recounted that \textit{``there were two nights that [child 1] and [child 2] had to remind me, \textit{`Mom, we have to do Misty,'} and I was like, \textit{`oh, yeah.'}'' }R1 also described getting reminders from their children, \textit{``the kids were curious about it, so they would also remind me like, \textit{`okay'}, you know, \textit{`go talk to the robot.  We're in bed.  Go talk to the robot.'}''}

\section{Discussion}
In this study, we explored the feasibility and effectiveness of using robots as interactive diaries and found that the robot was generally able to effectively collect the intended information. However, certain study scenarios and requirements may vary in their suitability for using an interactive diary robot. Below we describe scenarios where it may be advantageous to employ robots for diary studies and also discuss some limitations and extensions of the study.


\subsection{Diary Robots: Considerations for Applicability}
Reflecting on participants' diary entries and experiences, we identified the following factors to consider when deciding whether an interactive robot may be the preferred data collection instrument for a specific diary study.

\subsubsection{Descriptive or quantitative}
Our diary study consisted of six open-ended questions, and the robot thrived in producing descriptive and information-rich entries. Although the entries from the \textit{Audio} condition produced an overall similar level of information, they were potentially less focused compared to the entries created through the conversational robot. This pattern is consistent with previous studies that found that audio diaries are sometimes verbose and digressive~\cite{hyers2018diary}, which could result in more effort from the research team to analyze responses and a lower level of information richness on the core topics. Our interpretation is that participants perceived speaking to the robot to be a conversation akin to one with \textit{e.g.}, an experimenter, prompting them to focus on responding to the immediate question being asked.

\subsubsection{Complexity and sensitivity of the topic} 
Our interviews revealed that multiple factors unique to the robot have the potential to foster a sense of closeness (\textit{e.g.}, cuteness of the robot, and appropriate nonverbal expressions). A general consensus among all the participants was that the topic of the diary (\textit{i.e.}, bedtime routines) was not particularly personal or sensitive. For more complex or layered topics, an interactive robot diary has the potential to generate more in-depth insights.



\subsubsection{Importance of compliance}
Based on the interviews discussed in Section \ref{sec:compliance}, we speculate that the physical presence of the robot had a strong positive impact on compliance. Although it is unclear how this effect may translate to a long-term diary study, there is also potential in further utilizing the embodiment of the robot along with its other capabilities to encourage compliance. For example, the robot could use a blinking light to nudge the user to create an entry.

\subsubsection{Fixed location, or always on-the-go}
For all the potential benefits that a physical robot brings to the diary method, we trade off portability. If the study requires the creation of diary entries while the participant travels to different locations, relying solely on a diary robot system could be challenging. 

\subsection{Complementing the Diary Robot with Other Formats}
For studies that are challenging for a diary robot, such as those that require participants to provide a long series of numerical ratings, one possibility to take advantage of the strengths of the \textit{Diary Robot} could be a variation of the Diary-Interview Method~\cite{zimmerman1977diary}. Specifically, researchers could design their diary studies to include two components, where an alternative diary format complements the \textit{Diary Robot}. The complementary component could be designed to suit the particular need of the study (\textit{e.g.}, a questionnaire with a series of numerical ratings). The robot could then consume the collected data, and create synthesized diary entries by performing mini-interviews that combine and probe about the context of those earlier data collected.

\subsection{Limitations and Extensions}
Our study has a number of limitations that point to future research, which we outline below.



\subsubsection{Generalizability} Our study was carried out with participants with limited socioeconomic and demographic diversity. Future studies would need to be conducted with other population groups, and more participants, to assess the extent to which this method is applicable in other contexts, considering participant characteristics such as sex and preferences. For example, with older or less tech-savvy populations, interacting with a physical robot may be easier than navigating websites~\cite{robinson2014role}.
Our study was also relatively short, and focused on a single topic (\textit{i.e.}, bedtime routines). More studies are required to  examine how these results translate to longer-term diary studies and other topics.

\subsubsection{A Diary Study on the Robot Itself} Although in-the-wild HRI studies present a unique opportunity to use the robot deployed itself for the diary study, additional considerations are required when the robot is both the subject of the diary entries itself and the entity that facilitates the diary studies. Further research is needed to understand the nuances of such studies.

\subsubsection{Utilization of the Robot's Capabilities}
For comparability with existing diary formats, our robot's dialogue was very structured. \citet{wei2024leveraging} demonstrated the potential to utilize LLM-powered chatbots to collect self-reported data, and allowing the robot to perform diary studies through more free-form conversations would be an interesting extension. To this end, a future study that compares the diary robot system against a disembodied agent may also help single out the embodiment effect. Another a natural extension could be a variation of the media elicitation studies described by \citet{carter2005participants}, where the robot would capture photos and videos as it navigates within the environment, utilizing them as prompts and context reminders to create diary entries later.

\section{Conclusion}
In this study, we explored the use of a robot in diary studies as a data collection instrument. We developed the Diary Robot system, and carried out in-home studies to evaluate its effectiveness. We compared the diary entries created from these deployments to those created from existing diary methods (text- and audio-based diaries), and found that the Diary Robot was able to effectively elicit the intended information. Through post-study questionnaires and interviews, we further assessed the usability and applicability of an interactive robot diary. We discussed how it could be adapted to other scenarios, and noted some limitations and possible extensions.

\section*{Acknowledgment}
This work was supported by the National Science Foundation award \#2312354. We would like to thank Qiyao Yang for her excellent research assistance. Fig.~\ref{fig:teaser} used vector art assets by \texttt{storyset} from Freepik~\cite{freepik}. 

\balance
\bibliography{15_references}







\end{document}